\documentclass{article}

\usepackage{PRIMEarxiv}

\usepackage[utf8]{inputenc} 
\usepackage[T1]{fontenc}    
\usepackage{hyperref}       
\usepackage{url}            
\usepackage{booktabs}       
\usepackage{amsfonts}       
\usepackage{nicefrac}       
\usepackage{microtype}      
\usepackage{lipsum}
\usepackage{fancyhdr}       
\usepackage{graphicx}       

\usepackage{balance} 
\usepackage{graphics}
\usepackage{amsmath}
\usepackage{amssymb}
\usepackage{svg}
\usepackage{soul}
\usepackage{hyperref}
\usepackage{etoolbox}

\graphicspath{{media/}}     

\title{Reward-Sharing Relational Networks in Multi-Agent Reinforcement Learning as a Framework for Emergent Behavior
}

\author{
  Hossein Haeri, \ \  Reza Ahmadzadeh,  \ \  Kshitij Jerath \\ 
  University of Massachusetts Lowell \\
  \texttt{\{hossein\_haeri, reza\_ahmadzadeh, kshitij\_jerath\}@uml.edu} \\
}

\begin{document}
\maketitle

\begin{abstract}
In this work, we integrate `social' interactions into the MARL setup through a user-defined relational network and examine the effects of agent-agent relations on the rise of emergent behaviors.
Leveraging insights from sociology and neuroscience, our proposed framework models agent relationships using the notion of Reward-Sharing Relational Networks (RSRN), where network edge weights act as a measure of how much one agent is invested in the success of (or `cares about') another.
We construct relational rewards as a function of the RSRN interaction weights to collectively train the multi-agent system via a multi-agent reinforcement learning algorithm.
The performance of the system is tested for a 3-agent scenario with different relational network structures (e.g., self-interested, communitarian, and authoritarian networks).
Our results indicate that reward-sharing relational networks can significantly influence learned behaviors. We posit that RSRN can act as a framework where different relational networks produce distinct emergent behaviors, often analogous to the intuited sociological understanding of such networks.
\end{abstract}


\section{Introduction}
\label{sec:intro}

    Recent works in the field of Multi-Agent Reinforcement Learning (MARL) are taking the first steps towards developing a better understanding of interactions between artificially intelligent (AI) agents and their resulting emergent behaviors. While several interesting advances have been made thus far \cite{foerster2016learning, leibo2019autocurricula, lowe2017multi, vinyals2019grandmaster}, we still lack a broader framework that formulates the MARL problem in a manner that could subsequently generate a theory of emergent behaviors for a network of interacting AI agents.
    Such a theory of emergent behavior will require (a) a large number of AI agents with (b) complex interactions between the agents that subsequently produce (c) behaviors or dynamics that are discernible only at a global scale \cite{haeri2020thermodynamics, li2006survey, prigogine2018order, szabo2015formalization}. Within the context of MARL problems, this implies the need for a framework that can encapsulate complex agent interactions, learn associated policies, and measure global-scale dynamics (i.e. emergent behavior). 

        \begin{figure}[t]
            \centering
            {\includegraphics[trim={0 0.1em 0 0}, width=0.5\linewidth, clip]{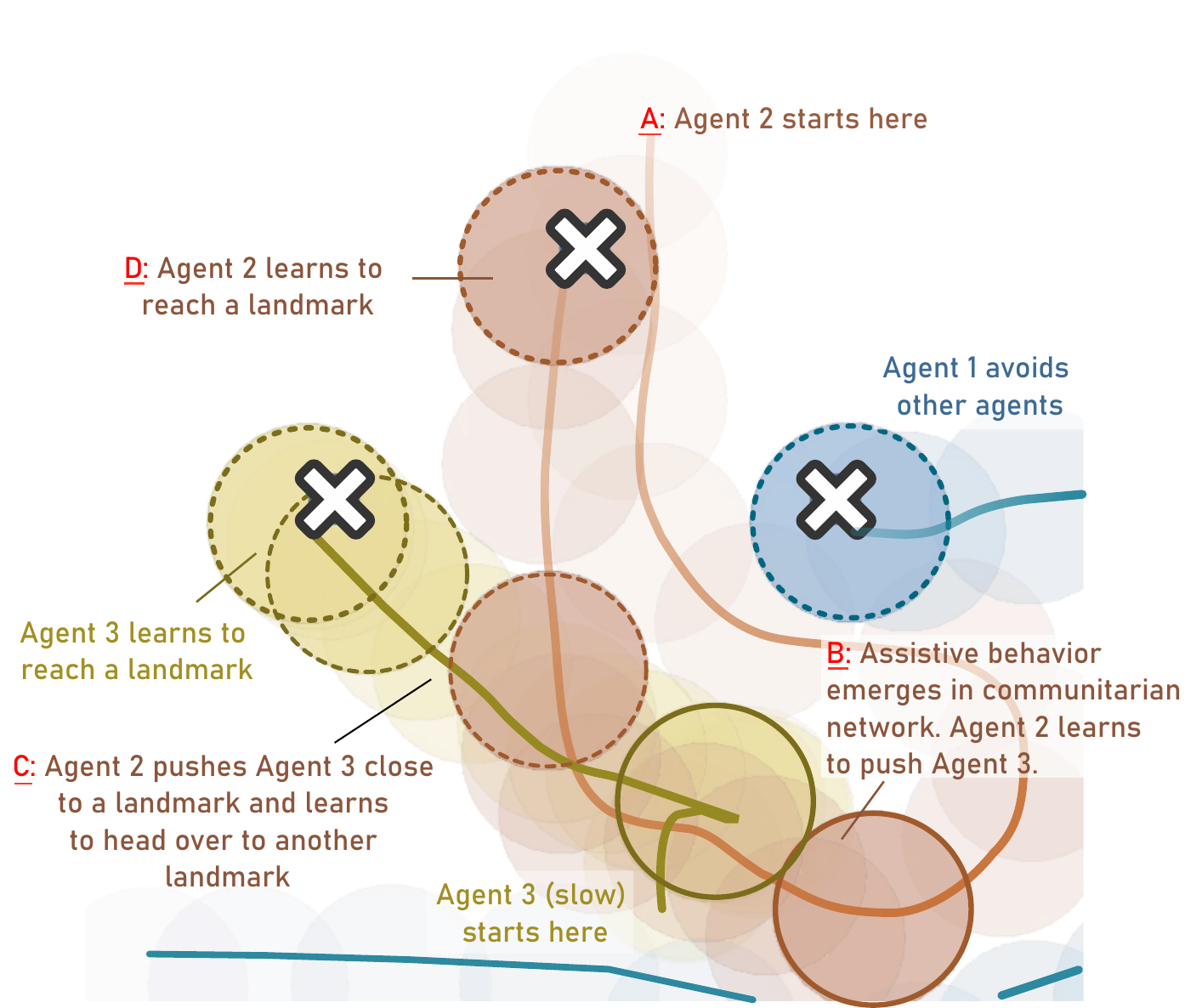}}
            \caption{Test episode for a trained communitarian (fully-connected) relational network system. Three agents try to reach three different landmarks (crosses), while agent 3's (yellow) movement is systematically hindered (slowed down). As expected of a communitarian social system, we observed the emergence of an assistive behavior where agent 2 (red) learns to push agent 3 towards another landmark before reaching its own landmark. 
            }%
            \label{fig:scenario}%
        \end{figure}    

\begin{figure}
        \centering
        {\includegraphics[width=0.6\columnwidth]{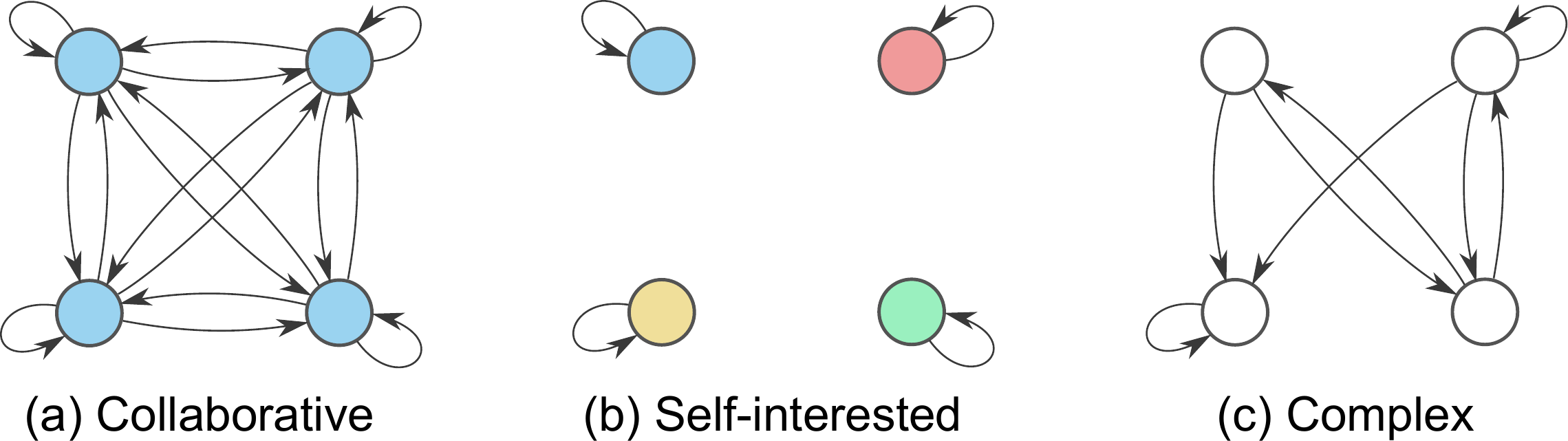}}
        \caption{In a decentralized MARL environment, agents co-evolve and learn simultaneously. (a) Collaborative agents  (b) Self-interested agents (c) Generalized complex network structure. Prior works have largely focused on problems associated with (a) and (b).
        }%
        \label{fig:schematic_networks}%
    \end{figure}

    Among the several ongoing MARL-related efforts in the research community, we are interested in approaches where learning happens not only as a result of interaction with the environment but also through complex `social' interactions among agents.  
    To be more specific, Training agents using a decentralized method \cite{baker2019emergent,omidshafiei2017deep} enables agents to co-evolve, therefore, learning becomes a \emph{collective} process. Moreover, while many state-of-the-art approaches primarily focus on either cooperative or competitive relationships, as illustrated in Figure \ref{fig:schematic_networks} (a) and (b), the designed MARL framework must go beyond and allow the learned behaviors to span an entire continuous spectrum of social behaviors, as illustrated in Figure \ref{fig:schematic_networks} (c). In many cases, such behaviors may only be possible via the implementation of distinct reward structures for disparate agents, as presented in Sections \ref{sec:methodology} and \ref{sec:results}.

    In this paper, we leverage findings from sociology and neuroscience to create a novel relational learning framework for MARL that can incorporate complex social interactions, which potentially increases our ability to scale and generate emergent behaviors in AI `societies' or `ecosystems'. We seek to demonstrate that this framework can produce an effective, systematic approach for implementing more complex relational network structures that address the needs of the MARL research community. There has been limited prior work on the role of \emph{relational} networks in collective learning in multi-agent systems \cite{Peysakhovich2018}. This is mainly because, though very common, complex networks arise in environments with more than a few agents. A consequence of working with several agents is that the state-action space suffers from the well-known `curse of dimensionality', which many existing policy optimization methods are unable to cope with.
    
    We also present the concept of Reward-Sharing Relational Networks (RSRN), implemented into the framework of Multi-Agent Reinforcement Learning (MARL) systems. Our RSRN-MARL framework helps determine how much each agent is invested in the success of (or `cares' about) other agents, quantified through their individual and relational reward signals. More importantly, in the long term, the notion of reward-sharing relational networks provides a framework and building blocks for creating and evaluating the performance of emergent behaviors in agent `societies'. We evaluate the effects of reward-sharing relational networks via two different reward sharing implementations and for six different network configurations. Each relational network is examined in a 3-agent MARL environment and analogies to sociological structures (such as self-interested, tribal, authoritarian, and communitarian networks) are provided. 
\section{Related Work}
\label{sec:related_work}
    Reward sharing in reinforcement learning is not in itself a new concept, with prior works on the topic dating back to the early 2000s \cite{bochi2003direct,mataric2001learning,miyazaki2001rationality}. One of the first instances of reward sharing in reinforcement learning discussed the idea using the notion of direct rewards (received as a result of an agent's own actions), and indirect rewards (received by other agents because of an agent's action) \cite{miyazaki2001rationality}. We use similar notions in later sections where we refer to individual (or direct) rewards and relational (or indirect) rewards in a network context. However, most related work in the domain has focused on game-theoretic approaches, which inherently limits the discussion of the problem to either a competitive \cite{kutschinski2003learning} or collaborative context \cite{oroojlooyjadid2019review, Leibo2017}. Other recent works have sought to move away from this dichotomous approach using the notion of mixed-motive games, even introducing scalar representations of prosocial propensity (such as social value orientation) \cite{mckee2020social}. However, a broader framework that eschews the dichotomous competitive vs. collaborative mindset for a more general relational network approach has largely been missing \cite{Peysakhovich2018}. Unlike these prior studies, the presented work borrows from the fields of sociology and neuroscience to provide a framework for emergent behavior by implementing reward sharing using an explicitly-defined `social' or relational network.

    There are also other recent efforts in the field of MARL that have studied multi-agent systems whose agents are able (or have the potential) to communicate with each other via a special communication channel to share information and coordinate with each other. \cite{foerster2016learning,zhang2019decentralized,kajic2020learning,lazaridou2020emergent}. These works have often been grouped collectively into the category of `networked' MARL, which may, at first glance, appear to have some similarity to the presented work. However, in these related works, agents usually learn the communication skills using a single team-level global reward function \cite{zhang2019decentralized}. Significant research efforts in this domain are directed towards solving, among others, the problem of communication-constrained distributed reinforcement learning.    
    
    Our study, however, is focused on the high-level ‘relationships’ between the agents. More specifically, the presented work focuses on capturing agent interactions in a `relational' graph that is also leveraged for reward-sharing and collective agent learning in the MARL framework. The weights of the graph edges between nodes (i.e. agents) represent how valuable another agent's success is to any given agent. This notion of reward sharing in a relational network is only tangentially related to the communication-focused research efforts performed under the umbrella of networked MARL. Section \ref{sec:implications} will provide additional details on this topic, with a discussion on different network structures and their impact on team rewards.

\section{Methodology}
\label{sec:methodology}
\begin{figure}
        \centering
        {\includegraphics[trim={0 0.1em 0 0},clip,width=0.7\columnwidth]{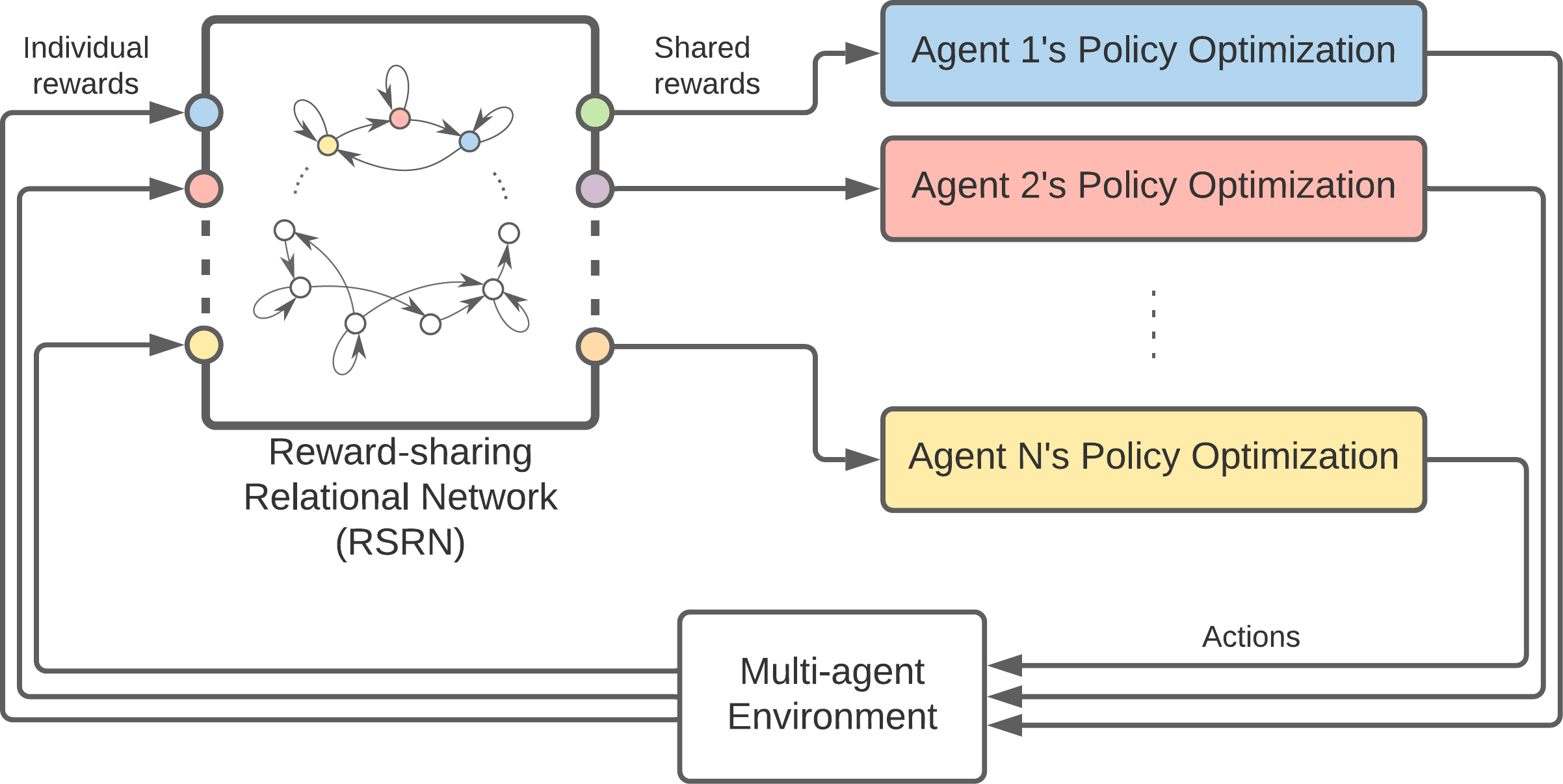}}
        \caption{Agents obtain individual rewards from the multi-agent environment. Shared rewards are evaluated using the Reward-Sharing Relational Network (RSRN) and depend on agent-agent relations. Different network structures generate different shared rewards, which are then used by the policy optimizer(s) and produce distinct emergent behaviors.
        }%
        \label{fig:diagram}%
    \end{figure}

       The primary innovation of the presented work is in the integration of relational networks with existing MARL methods which results in a novel framework that helps us understand how `social' interactions between agents generate emergent behaviors. These relational connections help determine how rewards obtained by one agent are shared with and drive the actions of another agent, and reflect which entities the agent `cares about'. Additional details on the sociological inspiration associated with this notion are provided in Section \ref{sec:implications}. As Figure \ref{fig:diagram} shows, our proposed framework adds a reward sharing block, called the Reward-Sharing Relational Network (RSRN), to a typical MARL flow diagram. The RSRN block takes a vector of individual agent rewards and a user-defined relational network as inputs and produces a vector of relational rewards for each agent as output. The output of shared relational rewards is used by the policy optimizer to generate a set of actions for every agent in the multi-agent environment.        
    
    \subsection{Problem Formulation}
\begin{figure}
    \centering
    \includegraphics[width=0.6\linewidth]{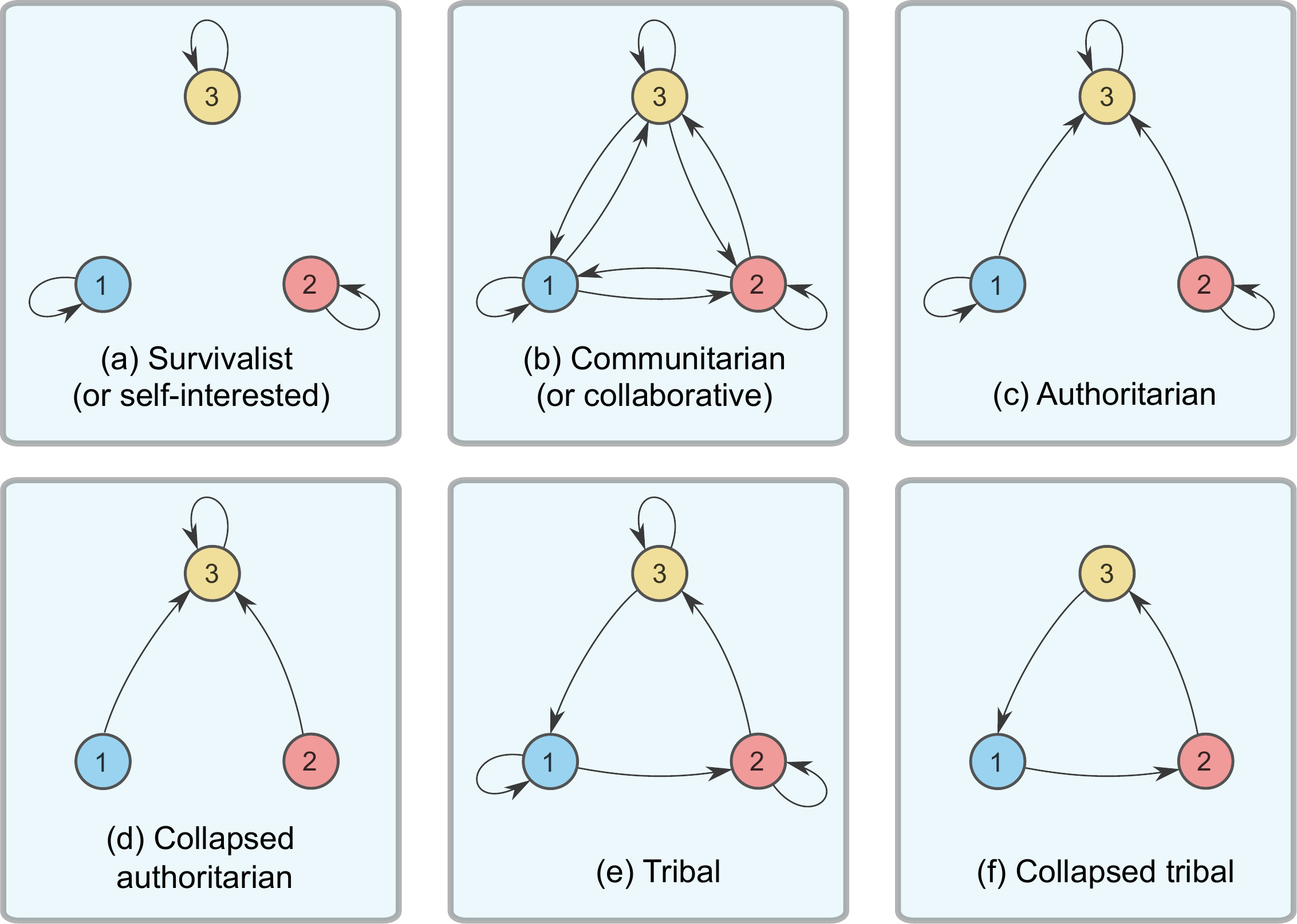}

    \caption{Examples of sociology-inspired relational network structures. Each of these relational networks is used as an RSRN in the 3-agent MARL system, and the resulting emergent behaviors are examined in Section \ref{sec:results}. An agent's actions may be governed by the rewards obtained by more than one agent (including itself). An arrow directed from a first agent to a second represents that the first agent's actions are governed by the rewards obtained by the second agent.}
    
    \label{fig:network_cofigs}%
    \end{figure}

    We define a Relationally Networked Decentralized Partially Observable Markov Decision Process (RN-Dec-POMDP) as a tuple $(\mathcal{S},\mathcal{A}, \allowbreak \mathcal{O}, \mathcal{T}, \mathcal{R},\mathcal{G})$, where $\mathcal{S}=\{S_1,S_2,...,S_N\}$, $\mathcal{A}=\{A_1,A_2,...,A_N\}$, $\mathcal{O}=\{O_1,O_2,...,O_N\}$, and $\mathcal{R}=\{R_1,R_2,...,R_N\}$ are the joint set of individual states, actions, observations, and rewards, respectively. The tuple element $\mathcal{G} = (V_{\mathcal{G}}, E_{\mathcal{G}}, W_{\mathcal{G}})$ represents the ordered collection of all agents as vertices in the set $V_{\mathcal{G}}$, all binary agent relations as directed edges in the set $E_{\mathcal{G}}$, and edge weights as $W_{\mathcal{G}}$. With reference to Figure \ref{fig:network_cofigs}, the presence of an edge between two agents represents that they are related. The direction of the edge from a first agent to a second agent represents that the actions of the first agent are driven by the rewards obtained by the second. This may be understood as the second agent \emph{sharing rewards} with the first agent, or that the first agent `cares about' the success of the related second agent. Consequently, the first agent is likely to learn policies that benefit the other agent. Of course, the relational network also allows for self-directed edges, so an agent's actions can also be driven by its own rewards. The nature of these relationships is encapsulated in the matrix $W_{\mathcal{G}}$ which denotes the weights associated with the edges. 
    
    At each time step $t$, each agent $i \in \{1,2,...,N\}$, first perceives the environment in the form of an observation $o_i \in O_i$, where $O_i$ is a set of its individual observations.
    Agent $i$ then takes action $a_i \in A_i$ based on its policy (in this case, deterministic) $\pi_i:O_i \rightarrow A_i$ and transitions (along with other agents) from to the next state according to the joint probabilistic transition function $\mathcal{T}:\mathcal{S}\times \mathcal{A} \times \mathcal{S}\ \rightarrow [0,1]$. Once an agent reaches its next state, it receives a reward $r_i$ according to its reward function $R_i:\mathcal{S} \times \mathcal{A} \rightarrow \mathbb{R}$.
    
    The relational network weights $W_{\mathcal{G}}$ can be expressed as a matrix whose elements $w_{i,j} \in \mathbb{R}$ indicate how much agent $i$ `cares about' the success of agent $j$, based on the individual reward obtained by agent $j$. Thus, actions of agent $i$ may be driven in part by the individual rewards of agent $j$, if these two agents are related, i.e. if $w_{i,j} \neq 0$.
    For example, a society with only self-interested agents, who are unrelated to and do not `care about' other agents, can be represented by the identity matrix. On the other hand, a communitarian society, that is highly collaborative and where relations exist between all agents, can be expressed by the all-ones matrix. Figure \ref{fig:network_cofigs} illustrates these two networks as well as other examples of RSRNs which will be discussed in section \ref{sec:implications}.  In general, due to the fact that the relational network includes directed edges, the weight matrix $W_\mathcal{G}$ is not necessarily symmetric. For example, in the tribal relational network shown in Figure \ref{fig:network_cofigs}(e), $w_{1,2} = 1$ (i.e. Agent 1 is invested in the success of Agent 2), but $w_{2,1} = 0$ (i.e. Agent 2 does not `care' about Agent 1).
    

    Integrating relational networks into the Dec-POMDP framework expands the research avenues associated with MARL to a spectrum of possible policies and behaviors depending on the definition of the weight matrix.
    Since most typical social systems are sparse, many weights $w_{i,j}$ are zero (e.g. small-world networks) \cite{amaral2000classes}.
    This key insight enables us to limit the reward sharing of an individual to only a few other agents. It also plays an important role in the tractability of finding the optimal policy, and perhaps an even greater role in future works seeking to scale the problem to larger agent populations.    
    
    \subsection{Scalarization of Individual Rewards}
    
    In our RSRN-MARL framework, agents
    try to maximize their long-term shared return $\bar{R}_i$ by accumulating the discounted 
    shared relational rewards $\bar{r}_{i}$ across a finite horizon $T$ as follows:        
        \begin{equation}
            \bar{R}_i = \mathbb{E}\left(\sum_{k=0}^{T}\gamma^k \bar{r}_{i,k}\right) = \mathbb{E}\left(\sum_{k=0}^{T}\gamma^k f(\mathbf{r}_k,\mathbf{w}_i)\right) 
        \end{equation}
        where $\bar{r}_{i}$ represents the shared relational reward of agent $i$, which incorporates the individual rewards of all related agents (and itself). Specifically, the scalarization function $f(\mathbf{r}_k,\mathbf{w}_i)$ maps all individual rewards $\mathbf{r}_k=[r_1,r_2,...,r_N]_k^\top$ at time step $k$ to a single shared relational reward value $\bar{r}_{i,k}$ for agent $i$ according to the agent-specific relational weight vector $\mathbf{w}_i=[w_{i,1}, w_{i,2}, ...,w_{i,N}]$, which determines how agent $i$ is related to other agents.
        
        It is possible to choose the scalarization function from several different alternatives, with the most commonly used choice often being a simple weighted sum across all the individual rewards, given by $\bar{r}_i = f_{\text{s}}(\mathbf{r},\mathbf{w}_i) = \sum_{j=1}^{N}w_{i,j} r_j \label{eq:WSM}$. However, we found that the Weighted Product Model (WPM), given by $
            \bar{r}_i = f_{\text{p}}(\mathbf{r},\mathbf{w}_i)=
            \prod_{j=1}^{N}r_j^{w_{i,j}}
            \label{eq:WPM} $, performed much better in our experiments while preserving the intuitive, qualitative relational between agents.
        Moreover, assuming that the rewards are positive, both  scalarization functions are strictly increasing for positive weights and are strictly decreasing for negative weights. Hence, they both satisfy the minimal assumption of scalarization functions \cite{ruadulescu2020multi}, i.e. the higher reward values map to higher scalarized values. 
        In this study, we used WPM scalarization due to the fact that it dramatically lowers the shared reward when even one of the individual rewards is close to zero. Therefore, trainers are strictly promoted to care equally about all the agents who are related with equal weights. 
    
    \subsection{Policy Optimization}
                     
        The presented work on relational networks does not rely on any specific policy optimization method. Depending on the application, we can use centralized or decentralized trainers. In centralized training, a single policy is optimized using all agents' action-observation feedback \cite{vinyals2019grandmaster}. Alternatively, one can use a set of single-agent policy optimizers to cope with this challenge. Although decentralized approaches are more general \cite{papoudakis2019dealing,foerster2017stabilising}, they have to deal with the non-stationarity and may not converge.
        Here, we briefly discuss one of these algorithms, Multi-agent Deep Deterministic Policy Gradient (MADDPG) \cite{lowe2017multi} that is an actor-critic method based on Deep Q-Networks (DQN) \cite{mnih2015human}. 
        In this algorithm, agent policies $\boldsymbol{\pi} = \{\pi_1, ..., \pi_N\}$ are represented using neural networks parameterized by $\boldsymbol{\theta}=\{\theta_1,...,\theta_N\}$.
        To optimize the policy, the gradient of the expected return for each agent can be written as $\nabla_{\theta_i} J(\theta_i) = \mathbb{E}(\nabla_{\theta_i}\log\pi_i(a_i|o_i)Q_i^\pi)$ \cite{lowe2017multi}.
        If we consider continuous deterministic policies $\mu_i$ parameterized by $\theta_i$, the gradient can be written as:
        
        \begin{equation}
            \nabla_{\theta_i} J(\mu_i) = \mathbb{E}(\nabla_{\theta_i} \mu_i(a_i|o_i) \nabla_{a_i}Q_i^{\mu})
        \end{equation}
        
        \noindent where $Q_i^\pi(\boldsymbol{o},\boldsymbol{a})$ is the centralized action-value function, $\boldsymbol{o}=\{o_1,...,o_N\}$ is the set of multi-agent observations, and $\boldsymbol{a}=\{a_1,...,a_N\}$ represents the set of multi-agent actions. The action-value function $Q_i^\pi(\boldsymbol{o},\boldsymbol{a})$ is then updated to minimize the loss function:
        \begin{equation}
            \mathcal{L}(\theta_i) =  \mathbb{E}\left[\left(Q_i^\pi-(\bar{r}_i+\gamma Q_i^{\pi'})\right)^2\right]
        \end{equation}
        where $\boldsymbol{\pi}'$ is a set of target policies with delayed parameters $\theta_i'$.
        Note that unlike MADDPG that uses the individual rewards $r_i$, we calculate the loss using agents' shared relational rewards $\bar{r}_i$. 

    \section{Structure of Reward-Sharing Relational Networks}
    \label{sec:implications}
    In this section, we will discuss how our work on reward-sharing relational networks (RSRN) is inspired in part by existing work on sociology \cite{Bales1950,Beck2012} and neuroscience \cite{Dunbar1992,jackson2005we}. 
    To form a framework for emergent behavior, we need to better understand how relational networks should be structured. We begin with two insights into the nature of learning in social systems comprised of primates (including humans) or other mammals \cite{DeWaal2007}.
    The first insight is that individual learning occurs in social settings \cite{Aharony2011}. 
    Specifically, it has been shown that learning amongst primates and other mammals is governed by whom the individuals can relate to. Individual learning is driven by actions ranging from simple mimicking to complex evaluative behaviors, all within the presence of other \emph{socially-related} individuals \cite{pezeshki2014s}. 
    The presented work on reward-sharing relational networks formalizes this notion within the context of MARL.
    
    The second insight borrows from neuroscience and indicates that learning individuals can connect or relate to a \emph{limited} number of their counterparts \cite{Dunbar1992}. This finding originates from studies of the size of the neocortex in primate brains and its relationship to the group size of the social systems these primates inhabit. 
    Building on these two insights, we propose that reward-sharing and learning in a multi-agent system should: (a) be governed by whom an agent is related to, and (b) be limited to a small number of relations to obtain different types of emergent behaviors. 
    The relational networks thus generated are used for training the agents, and may potentially generate distinct emergent behaviors depending on the structure of the relational network. We evaluate the proposed framework via simulations of a 3-agent system in different environments and the results are presented in Section \ref{sec:results}.

    \subsection{Implications of Relational Network Structures}
    The structure of relational networks that exist in social systems (such as the human society) have deep implications on the policies learned and emergent behaviors observed in such systems \cite{Beck2012,Aharony2011}. For example, strictly hierarchical relational networks produce authoritarian emergent behaviors, where individuals are significantly driven by the states, actions, and performance of the individuals higher up in the hierarchy \cite{aronoff1971motivational}. 
    One may intuitively expect to observe similar emergent behaviors in other similarly structured `social' system of AI agents, such as in our RSRN-MARL framework.
    
    To provide the reader with a better description, we now discuss other examples of relational networks with reference to Figure \ref{fig:network_cofigs}. 
    As seen in Figure \ref{fig:network_cofigs}(a), \emph{survivalist} relational networks consist of individuals that are primarily self-interested and focused on self-preservation. In a MARL system, this may translate into keeping one's reward for oneself, so that an agent's actions are governed only by the reward it gets for itself. The associated relational network weight matrix $W_G$ is, in its simplest form, an identity matrix, with no relationships with other individuals. Prior works may model this behavior via non-cooperative games. On the contrary, \emph{communitarian} relational networks are highly collaborative in nature, and each individual is related to every other individual and `cares' about them, as shown in Figure \ref{fig:network_cofigs}(b). In a MARL system, this may be implemented as reward sharing with all agents, such that the policies learned by an agent are a function of the rewards of \emph{all} other agents, and not just its own rewards. Prior works may model this behavior to some extent as collaborative games \cite{lowe2017multi}.
    
    To the best of our knowledge, the other relational networks do not have distinct, formalized counterparts in prior MARL literature, and several other network configurations not shown may also be evaluated. Figure \ref{fig:network_cofigs}(c) represents an \emph{authoritarian} relational network which is defined by its hierarchical elements \cite{Beck2012}. Specifically, in the authoritarian relational network, all individuals in the social system are invested in their own success, but their actions are also driven by the need to ensure the success of those above them in the hierarchy. E.g., in the context of a 3-agent MARL problem shown in Figure \ref{fig:network_cofigs}(c), the directed edge from agent 1 to the `higher' agent 3 indicates that agent 1's actions are governed by agent 3's rewards. However, there is no directed edge from agent 3 to the `lower' agent 1, as agent 3's policies are not driven by the individual rewards of agent 1.
    A devolved version of the network in Figure \ref{fig:network_cofigs}(c) may represent a \emph{collapsed authoritarian} relational network (Figure \ref{fig:network_cofigs}(d)), where the hierarchy is maintained, but exists solely for the purpose of benefiting the individuals higher in the hierarchy. 
    
    Similarly, Figure \ref{fig:network_cofigs}(e) represents a \emph{tribal} relational network, where along with being driven by one's own success, an individual is also driven by the success of specific other individuals with whom it shares a familial relationship. In the MARL context, and especially in the 3-agent network, we represent this as directed edges from one agent to the next. These tribal relational networks will be easier to visualize and comprehend with larger multi-agent systems. A devolved version of the network in Figure \ref{fig:network_cofigs}(e) may represent a \emph{collapsed} or \emph{poorly scaffolded tribal} relational network as shown in Figure \ref{fig:network_cofigs}(f). In this relational network, individuals only care about the success of their familial relations and are not concerned with their own success. 
    

\section{Simulation and Results}
\label{sec:results}
    In this section, we implement an RSRN on a 3-agent MARL environment and evaluate the performance of the agents under different network structures. The environment is simulated using the Multi-agent Particle Environment (MPE) \cite{lowe2017multi} where agents are simulated as physical elastic objects. 
    Both state and action spaces are continuous and agents have been trained using our framework by integrating the Relational Network and Multi-Agent Deep Deterministic Policy Gradient (MADDPG) policy optimization algorithm \cite{lowe2017multi}. 
    We used 2 fully-connected 64-unite layers for policy networks which are optimized with the learning rate of 0.01, batch-size of 2048, and the discount ratio of 0.99.

         As shown in Figure \ref{fig:scenario},
     We consider the scenario where three agents try to reach three unlabeled landmarks, i.e. they get rewarded if they reach any landmark. To make the multi-agent environment more complex and create opportunities to observe emergent behaviors arise, we limit the mobility of one of the agents. Specifically, agent 3 is hindered systematically so it cannot move as fast as other agents, and may be unable to reach a landmark within a single episode.

    To evaluate the collective, emergent behavior of the agents, we examine 6 different relational network configurations as shown in Figure \ref{fig:network_cofigs}. The individual performance of each agent $i$ is measured through its individual reward $r_i$, which is $e^{-d_{i}^2/\sigma^2}$ if $d_{i}$ is less than 0.2 and is zero otherwise,
    where $d_i$ denotes the Euclidean distance to the closest landmark and constant $\sigma^2$ is set to $0.1$. It should be noted that the landmark locations are fixed in these simulations but agents are initialized at random positions. 
    
    After training agents, we run test episodes with the learned policies to calculate their mean individual and shared relational rewards which correspond to their individual and social performances, respectively. Figure \ref{fig:bars_WPM}(a) shows individual rewards which determine how successful each agent reaches the landmarks. Similarly, Figure \ref{fig:bars_WPM}(b) shows shared relational rewards which indicate the performance of the related agents. Additional discussion related to this topic can be found in Section \ref{subsec:examining_emergent_behavior}.        
    These figures and associated videos\footnote{
    Videos are available at  \hyperlink{https://sites.google.com/view/marl-rsrn}{\normalfont {\texttt {sites.google.com/view/marl-rsrn}}}
    } help reveal insights into how different relational networks produce distinct emergent behaviors, as well as the performance of individual agents.

    \subsection{Examining Emergent Behaviors in Relational Networks}
        \label{subsec:examining_emergent_behavior}
        In the following paragraphs, we will briefly discuss the emergent behaviors for each network configuration.
        
        \begin{figure}[t]
        \centering
        {{\includegraphics[width=.9\linewidth]{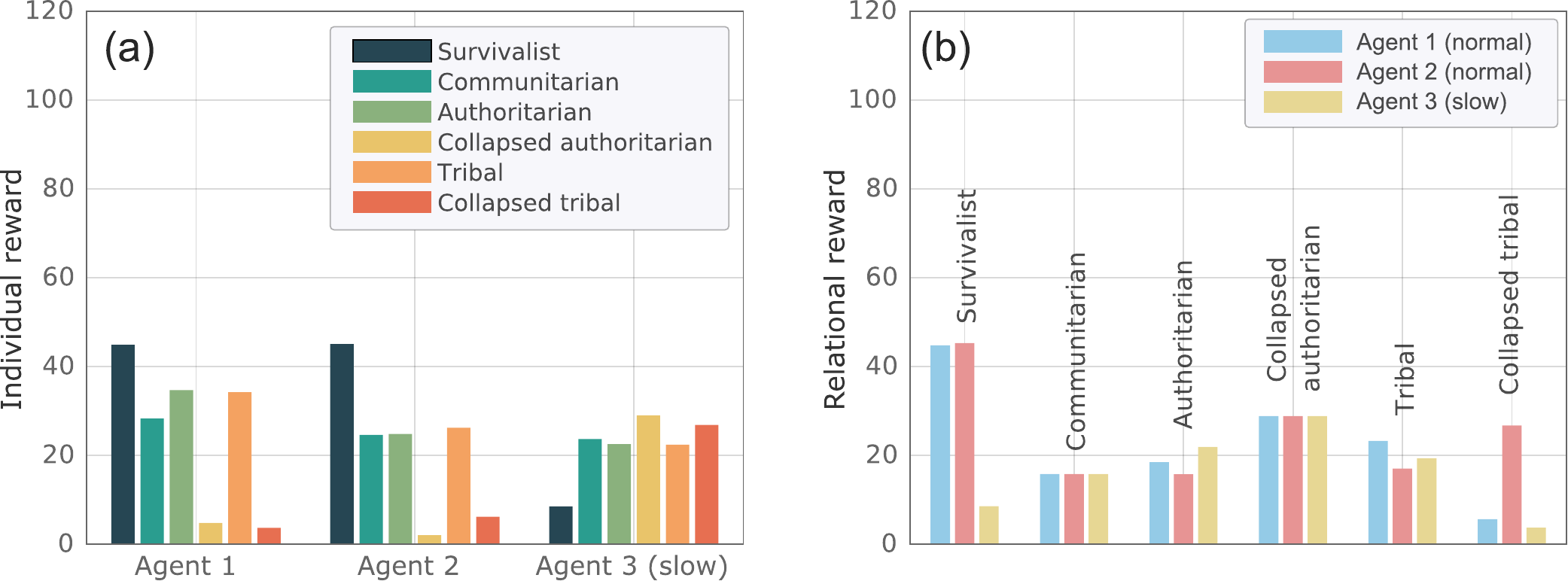}}}
        \caption{(a) Individual rewards and (b) relational rewards using WPM scalarization. Here, all rewards are averaged across 5000 test episodes, after training over 500,000 episodes.}
        \label{fig:bars_WPM}%
        \end{figure}       
    
        \noindent\emph{\underline{Survivalist} (or self-interested)}: In this network configuration, the shared relational reward for each agent is the same as its own individual reward, since no relations exist between agents. 
        Agents 1 and 2 easily learn policies that generate significant individual rewards. However, the constrained agent 3 is unable to collect meaningful rewards. In fact,  as is evident from Figure 6(a), agent 3 even loses some of its initial randomly collected reward because the other agents learn to serve their own interest and push agent 3 away from a closer landmark if needed. As shown in the tests results in \ref{fig:bars_WPM}(a), agent 3 fares much poorly than the other agents in this survivalist relational network.
        
        \noindent\emph{\underline{Communitarian} (or collaborative)}: In this setup, $w_{i,j} = 1, \: \forall \: i,j$. Consequently, all agents receive identical rewards, and they form a homogeneous team where no one has any reward privileges over others. Figure \ref{fig:bars_WPM}(a) indicates that all three agents receive similar individual rewards in the test phase. More importantly, this is a direct result of the identical shared relational rewards received by the three agents, as shown in Figure \ref{fig:bars_WPM}(b). While agents 1 and 2 receive lower rewards as compared to the survivalist relational network, each agent in a communitarian network is taken care of.

\begin{figure*}[h]
    \centering
    \includegraphics[width=\linewidth]{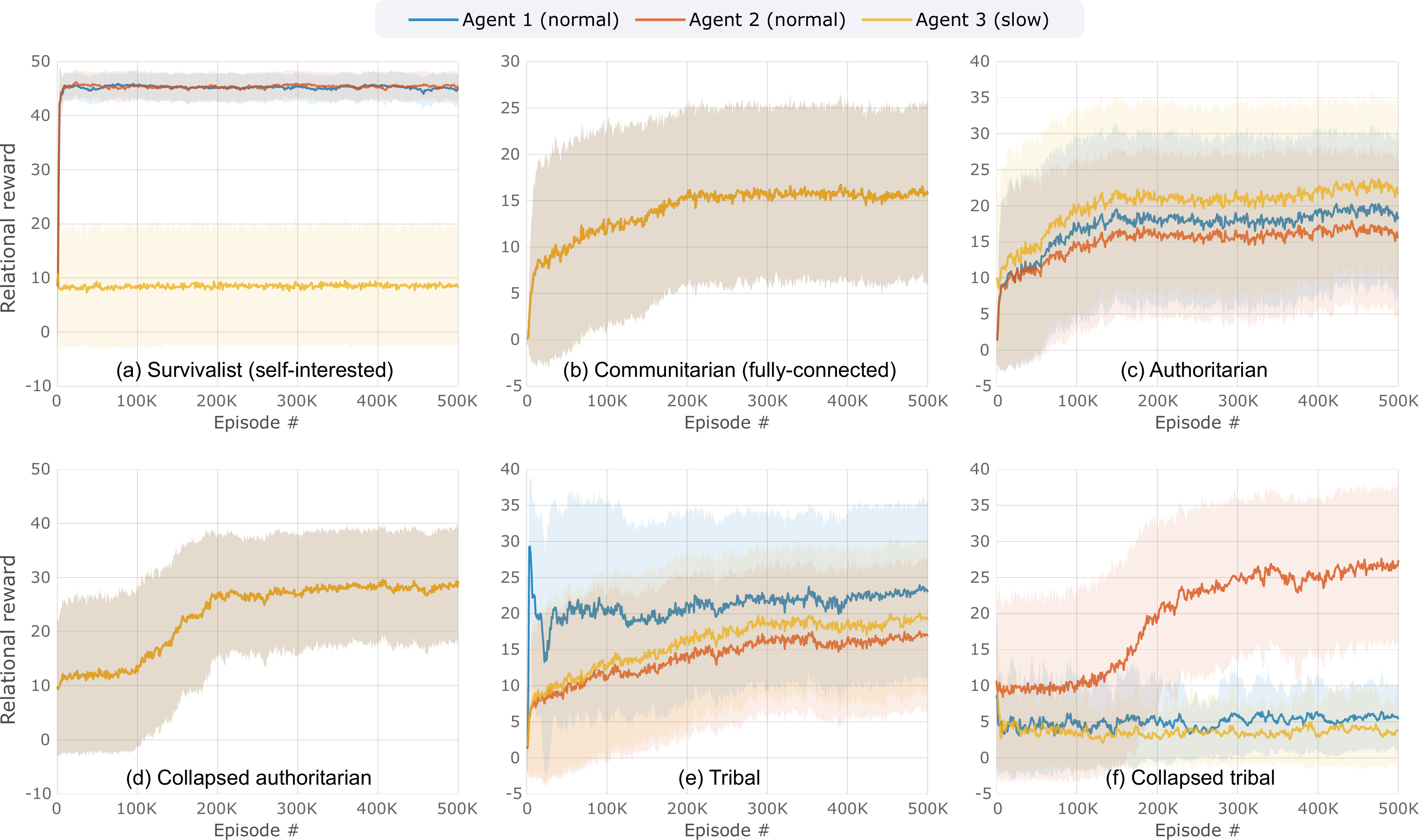}
    \caption{The relational reward of agents as they are being co-trained with different RSRNs.}%
    \label{fig:rewards_WPM}%
\end{figure*}
        
    \noindent\emph{\underline{Authoritarian}}: In this relational network (RN), agents 1 and 2 weigh their own rewards as much as they weigh the rewards received by the slow agent 3 (`the monarch'). The emergent behavior observed here includes agents 1 and/or 2 assisting agent 3 to the landmark before reaching other landmarks themselves. 
    
    \noindent\emph{\underline{Collapsed Authoritarian}}: This relational network demonstrates even more interesting emergent behaviors than the authoritarian RN. As a potentially devolved version of the authoritarian network, this RN demonstrates behaviors associated with extreme subservience to the rewards of agent 3. In other words, agents 1 and 2 learn policies that only benefit agent 3 (`the dictator'). These agents receive minimal individual rewards for themselves, since $w_{1,1} = w_{2,2} = 0$ (Figure \ref{fig:bars_WPM}(a)), but significant relational rewards, since $w_{1,3} = w_{2,3} = w_{3,3} = 1$ (Figure \ref{fig:bars_WPM}(b)). In video evidence, agents 1 and 2 are seen to assist agent 3 to the landmark and either continue to loiter around agent 3 to make minor adjustments that may increase its reward, or eventually wander away in the environment.
        
    \noindent\emph{\underline{Tribal}}: Each agent $i$ in this configuration cares equally about itself and a specific 'friend' or familial agent $j$, though this is not reciprocated by agent $j$. In the specific configuration considered in \ref{fig:network_cofigs}(e), the relationship is cyclic, so each agent cares about the next. This produces a sustainable tribal relationship, as evinced from the comparable individual and relational rewards across all three agents seen for tribal RN in Figure \ref{fig:bars_WPM}. 
        
    \noindent\emph{\underline{Collapsed Tribal}}: In this relational network, we observe a devolved tribal or familial relationship, where individual agents no longer learn policies associated with their own rewards but only of their related agent's rewards.    
    Agent 1 performs poorly with low individual and relational rewards, since only agent 3 `cares' about it, and agent 3 is systematically hindered or slow. Hence, agent 3 is unable to learn requisite actions that would increase the rewards of agent 1. Agent 2, on the other hand, can move fast and learn optimal policies that increase individual rewards for agent 3, and consequently the shared relational rewards for agent 2. Thus, we observe from Figure \ref{fig:bars_WPM} that agent 3's individual reward and agent 2's relational reward are significant, respectively. Moreover, agent 3's individual rewards are higher than those received in a tribal RN, but this is at the cost of severely reduced performance of the other agents. 
    Video evidence indicates that agent 2 hovers close to agent 3, even after the latter has reached a landmark. While it was intuitively expected that agent 1 would intervene and push agent 2 towards another landmark after agent 3 had been assisted, we did not observe such behavior. One possible reason could be that agents 1 and 2 are equally fast, and agent 1 is unable to learn to push agent 2 in any meaningful way.

\vspace{-0.2cm}

\section{Conclusions and Future Work}
Driven by insights from sociology and neuroscience, we have proposed a Reward-Sharing Relational Network (RSRN) multi-agent reinforcement learning framework that formalizes social learning by determining how agents `care about' each other. The agents learn policies that maximize their shared relational rewards, which are constructed according to the individual rewards of all related agents. This framework has the potential to act as a foundation for studying agent interactions and the generation of emergent behaviors in societies or ecosystems of learning agents.

We have demonstrated the novelty and effectiveness of this approach using a 3-agent MARL system with various relational network configurations, where agents learn to reach fixed landmarks. As Figure \ref{fig:bars_WPM}(a) showed, different relational networks result in distinct learned behaviors of finding landmarks and individual agent performances. The learned policies often include emergent behaviors not explicitly defined in the problem formulation. E.g., agents in a \emph{survivalist} relational networks quickly learned to go to a landmark, even pushing slower agents out of the way, though no such behavior was included in the problem formulation. On the other hand, agents trained with a \emph{communitarian} relational network adapt and learn to distribute their tasks according to their capabilities. The system exhibits emergent behavior as fast-moving agents learn to assist the slow-moving agent towards the landmark, before finding a landmark of their own. Similar emergent behaviors manifest in \emph{authoritarian} and \emph{tribal} relational networks as well. E.g., in authoritarian relational networks, fast-moving agents lower in the hierarchy assist slow-moving agents higher in the hierarchy. However, these services are not afforded to slow-moving agents if they exist lower in the authoritarian hierarchy. Collapsed authoritarian and tribal networks also demonstrate some extreme behaviors where some agents are subservient to the rewards received by other agents and continue to hover around slow-moving agents, even after they have been assisted towards a landmark.

Among the several challenges associated with studying emergent behaviors in the MARL problems, a key issue holding the field back is the apparent intractability and difficulty in scaling the problem to a large number of agents, which happens to be a key requirement to study emergent dynamics of interacting AI agents. Our future works will be directed towards leveraging the Reward-Sharing Relational Networks in Multi-agent Reinforcement Learning problems to specifically tackle the challenge of scalability. Potential approaches to expand upon the presented work include (a) the expansion of RSRNs to dynamic or evolving relational networks (unlike the current work, which assumed static network configurations), (b) the use of fractional or negative weights to model less-than-collaborative or hostile relationships, respectively, 
and (c) estimating the rewards obtained by related individuals. We expect these advances to provide the first steps towards building a theory of emergent behaviors in interacting AI agents.

\vspace{-0.3cm}
\section*{Acknowledgments}
\vspace{-0.3cm}
The authors would like to acknowledge their discussions with and insights provided by Dr. Charles Pezeshki at Washington State University, regarding the sociological analogues associated with specific reward-sharing relational networks.
\vspace{-0.3cm}
\bibliographystyle{unsrt}  
\bibliography{main.bbl}  

\vspace{-0.3cm}
\end{document}